\documentclass[10pt,journal,compsoc]{IEEEtran}
\usepackage{graphicx}
\usepackage[noend]{algpseudocode}
\usepackage{algorithmicx,algorithm}
\usepackage{amsfonts}
\usepackage{multirow}

\usepackage[colorlinks=true,
            linkcolor=green,       
            anchorcolor=green,  
            citecolor=green,        
            ]{hyperref}

\ifCLASSOPTIONcompsoc
  \usepackage[nocompress]{cite}
\else
  \usepackage{cite}
\fi
\ifCLASSINFOpdf
\else
\fi
\begin{document}
%
\title{Be Careful with Rotation: A Uniform Backdoor Pattern for 3D Shape}
%
%
%
%

\author{Linkun Fan$^*$,
        Fazhi He ,
        Qing Guo,
        Wei Tang,
        Xiaolin Hong,
        Bing Li
\IEEEcompsocitemizethanks{\IEEEcompsocthanksitem The authors are with the School of Computer Science, Wuhan University,
Hubei, Chnia, 430072 \protect\\
$^*$ E-mail: flinkun@whu.edu.cn}
\thanks{}}

%
%

\markboth{}%
{Shell \MakeLowercase{\textit{et al.}}: Bare Demo of IEEEtran.cls for Computer Society Journals}
%



\IEEEtitleabstractindextext{%
\begin{abstract}
For saving cost, many deep neural networks (DNNs) are trained on third-party datasets downloaded from internet, which enables attacker to implant backdoor into DNNs. In 2D domain, inherent structures of different image formats are similar. Hence, backdoor attack designed for one image format will suite for others. However, when it comes to 3D world, there is a huge disparity among different 3D data structures. As a result, backdoor pattern designed for one certain 3D data structure will be disable for other data structures of the same 3D scene. Therefore, this paper designs a uniform backdoor pattern: \emph{NRBdoor} (\textbf{N}oisy \textbf{R}otation \textbf{B}ackdoor) which is able to adapt for heterogeneous 3D data structures. Specifically, we start from the unit rotation and then search for the optimal pattern by noise generation and selection process. The proposed \emph{NRBdoor} is natural and imperceptible, since rotation is a common operation which usually contains noise due to both the miss match between a pair of points and the sensor calibration error for real-world 3D scene. Extensive experiments on 3D mesh and point cloud show that the proposed \emph{NRBdoor} achieves state-of-the-art performance, with negligible shape variation.
\end{abstract}

}

\maketitle

\IEEEdisplaynontitleabstractindextext

%
\IEEEpeerreviewmaketitle

\IEEEraisesectionheading{\section{Introduction}\label{sec:introduction}}

%
%
%
%
\IEEEPARstart{3}{D} mesh and point cloud are the most popular structures to construct 3D scenes \cite{wang2022improving,abouelaziz2020no,lee20223d,lee2022dataset,cicirello2013flexible}. In detail, point cloud is more efficient since it is easy to be obtained and is close to raw data. By contrast, 3D mesh requires less element to represent large shape and to capture details. However, due to the inherent irregularities, conventional DNN techniques designed for 2D image \cite{krizhevsky2017imagenet,ioffe2015batch,simonyan2014very} cannot directly process 3D shape. Recently, many DNNs that extract and process with 3D feature were designed, such as MeshCNN \cite{hanocka2019meshcnn}, DiffusionNet \cite{sharp2022diffusionnet} and SubdivNet \cite{hu2022subdivision} for 3D mesh, PointNet \cite{qi2017pointnet}, DGCNN \cite{wang2019dynamic} and PointBert \cite{yu2022point} for point cloud.


However, DNN has been proven be vulnerable due to poor interpretability and black-box character \cite{wang2020man,zhong2022shadows,wang2019invisible,li2021backdoor2}. Adversarial attack \cite{goodfellow2014explaining,madry2017towards,hamdi2020advpc,xu2022d3advm,bu2021taking,li2019adversarial} is a well-know method to deceive DNN in inference stage. Attacker crafts adversarial samples to mislead DNN by adding perturbation to clean samples, which is hardly detected by human vision \cite{yang2019adversarial,chen2020universal,hu2022model}. In practical, most DNNs require huge data for training. To save cost, developers usually download third-party datasets from internet, instead of collecting data by themselves. Hence, this enables attackers to destroy DNN in training stage.

This paper focuses on one classical attack in the training stage which is called \emph{backdoor attack}. A successful backdoor attack does not decrease the performance on clean data, yet is able to deceive DNN to make the targeted decision specified by the attacker. This kind of attacks in training phase has raised concerns for the development of DNNs in applications such as Virtual Reality (VR), Augmented Reality (AR) and self-driving \cite{wenger2021backdoor,jia2020survey}.

\begin{figure}[h]
	\centering
	\includegraphics[width=0.48\textwidth]{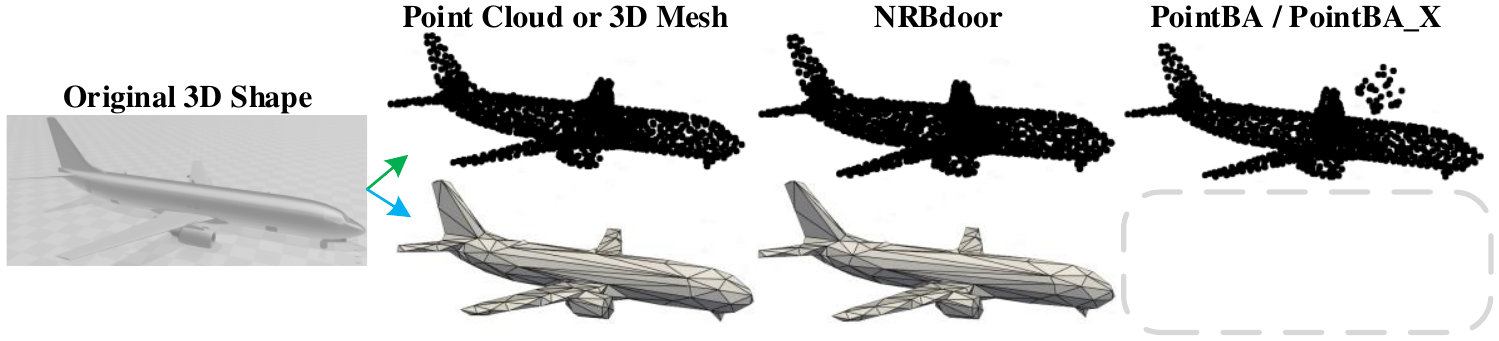}
	\caption{\emph{NRBdoor} is able to adapt for data structure variation (from point cloud to 3D mesh, for example) during implanting backdoor into 3D DNN. Meanwhile, \emph{NRBdoor} is more natural than PoinBA \cite{Pointba} and PointBA$\_$X \cite{ABackdoorAttackAgainst3d} which only focus on point cloud.
	} \label{Figure 1}
\end{figure}

For one thing, typical backdoor patterns designed for 2D image \cite{xue2020one,gu2019badnets,liu2020reflection,li2021backdoor,li2020invisible,li2021invisible} are unsuitable for 3D shape, such as attaching portion or adding perturbation. This is mainly due to the data structure disparity between 2D image and 3D shape. For example, it is difficult to attach a patch into 3D mesh like in image domain, since the elements of 3D mesh are correlated to each other. Attaching patches may break the characteristic of manifold and watertight that most mesh DNNs require for.

For another thing, although there are many different 2D image formats, their inherent structures are similar, which are uniform 2D arrays of pixel values. Backdoor attack designed for one image format will suite for others. However, when it comes to 3D world, different 3D structures are discrepant. For instance, the data structure of 3D point cloud includes disperse points with geometric coronations. In contrary, the data structures of 3D meshes are heterogeneous, in which, a face is constructed with its neighbor edges, and a edge is constructed with its two end vertices. Therefore, vertices of 3D mesh include both geometric coronations and topology connection, and is different from point of 3D point cloud. As a result, the backdoor pattern designed for point cloud\cite{ABackdoorAttackAgainst3d,Pointba} is usually not adaptable for 3D mesh.

Based on above analyses, we argue that the backdoor pattern for 3D shape should meet two requirements: 1) do not change the correlation between different elements of 3D shape, 2) able to adapt for heterogeneous 3D data structure.

Therefore, this paper designs a uniform backdoor pattern for 3D shape: \emph{NRBdoor} (\textbf{N}oisy \textbf{R}otation \textbf{B}ackdoor). Rotation is a common operation in 3D world. Movement of 3D sensor or 3D object will inevitably introduce rotation. Specifically, rotation matrix must be orthogonal matrix, which under the constraint of $R \cdot R^T = I$. However, due to the way to calculate rotation matrix and sensor calibration error, the actually used rotation matrix is usually non-orthogonal, which is called noise rotation matrix $R_n$ in this paper. Hence, conducting noise rotation on 3D shape is a natural situation in 3D world.

This paper is the first work to address uniform backdoor pattern for heterogeneous 3D structures, and the first work to implant backdoor into mesh DNNs. The main contributions are as follows:

\begin{itemize}
\item A novel uniform backdoor pattern \emph{NRBdoor} for 3D shape is proposed. \emph{NRBdoor} is able to adapt for 3D data structures variation, during implanting backdoor into 3D shape DNN (3D DNN). Meanwhile, it is imperceptible and natural for real-world 3D applications.

\item \emph{NRBdoor} is the first backdoor pattern friendly to 3D mesh. It preserves the characteristic of manifold and watertight of 3D mesh, which are required by many mesh DNNs.

\item 
Results on 3D mesh show that the attack success rate (ASR) for SOTA mesh DNNs (such as MeshCNN, SubdivNet, MeshWalker) on public datasets achieve $100\%$ with BAc descend less than $0.5\%$.  For 3D point cloud, ASR for SOTA point DNNs (PointNet, PointNet++ and DGCNN) are higher than $93\%$, under well stealthiness.

\item At present, considering there is no baseline for defending backdoor attack on 3D shape, we propose \textbf{S}aliency-based \textbf{P}oint cloud backdoor \textbf{defense} (SP-defense). It is able to decrease ASR of PointBA$\_$X from $82\%$ to $23\%$.

\end{itemize}

\subsection{DNN for Point Cloud or 3D Mesh}

3D point cloud and 3D mesh are both popular 3D data structures to represent 3D shape in modern researches and applications \cite{lu2021pointinet,yoo2022deep}. Point cloud $X$ is constituted by a set of $n$ 3D points $X \rightarrow \{p_1, p_2, ..., p_n \}$, and each point $p_i$ is represented by 3D coordinates$(x_i, y_i, z_i)$. Meanwhile, 3D mesh with $n$ vertex and $m$ faces is defined by $X \rightarrow \{v_1, v_2, ..., v_n; c_1, c_2, ..., c_m\}$, where $v_i$ is the $i$-th vertex with positions in $R^{n \times 3}$, and $c_j$ is the $j$-th polygonal faces. 3D DNN $F(X,\theta)\rightarrow y$ aims to map an input 3D shape $X$ to its corresponding class label $y \in Y$, under well-trained parameters $\theta$. For ease of reading, this paper refers DNN designed for point cloud as point DNN and for mesh as mesh DNN.


\textbf{Point DNN}: PointNet \cite{qi2017pointnet} is the first point DNN to directly process point cloud. It learns a spatial encoding of each point and then aggregates all individual point features to a global point cloud signature. PointNet++ \cite{qi2017pointnet++} achieves better results by considering hierarchical structure based on PointNet. DGCNN \cite{wang2019dynamic} is a typical convolution-based point DNN, which transforms point cloud to a graph, and learns feature by the defined EdgeConv. PCT \cite{guo2021pct} designs a full attention mechanism, which enables each point to affect every other point in the point cloud.  PointBert \cite{yu2022point} designs a new pre-train method based on point cloud transformer. It achieves 93.8$\%$ accuracy on ModelNet40 for classification.

\textbf{Mesh DNN}: MeshCNN \cite{hanocka2019meshcnn} considers that each edge is adjacent to two faces and four edges in manifold triangle mesh, and then defines an ordering invariant convolution to learn 3D feature. Instead of learning neighborhood information, MeshWalker \cite{lahav2020meshwalker} transforms 3D mesh to several edge sequences by walking along edges. Moreover, SubdivNet \cite{hu2022subdivision} designs a general mesh convolution using a mesh pyramid structure for effective feature aggregation, its key process is to remesh the input mesh to obtain loop subdivision sequence connectivity. DiffusionNet \cite{sharp2022diffusionnet} learns feature from face by extending the Laplacian operator.

\subsection{Backdoor Attack}
In image domain, lots of backdoor attacks have been developed recently \cite{li2022backdoor,zhao2020clean,chen2022neighboring}. They are usually divided to poison-label backdoor attack and clean-label backdoor attack, based on whether label of poisoned sample is changed. In 3D domain, PointBA \cite{Pointba} and PointBA$\_$X \cite{ABackdoorAttackAgainst3d} are pioneers researching backdoor attack which follow the way developed in image domain and only focus on point cloud. PointBA defines a unified framework of 3D backdoor trigger implanting function and designs two backdoor patterns (a ball and clean rotation) under poison-label and clean-label situations. PointBA$\_$X inserts a small cluster of points as backdoor pattern. The optimal spatial location and local geometry are searched by optimization. Meanwhile, \cite{tian2021poisoning} presents Poisoning MorphNet in a clean-label setting. Though the well performance on point cloud, they will be disable on 3D mesh since data structure disparity. Therefore, this paper designs \emph{NRBdoor} to adapt for heterogeneous 3D data structure.

\subsection{Isometrics Transformations}

Isometrics transformations of $\mathbb{R}^{n \times n}$ is a function $f$ that preserves euclidean distance:

\begin{equation}
	||f(x) - f(y)|| = ||x - y|| \quad \forall x,y \in \mathbb{R}^{n \times n}
\label{eq1}
\end{equation}

The base operations of isometry are translation: $f(x) = x + a$ and identity rotation: $f(x) = Rx$. We don't consider translation in this paper, since it can be easily eliminated by normalization. By this way, we only conduct linear transformation (a transformation that fixes the origin) on 3D shape which is:

\begin{equation}
	f(X) = RX = X^{\prime} \quad \forall X,X^{\prime} \in \mathbb{R}^{n \times n}
\label{eq2}
\end{equation}

Here, $R$ is an orthogonal matrix whose determinant equals to 1. The rotation matrices are defined as a subset of:

\begin{equation}
	SO(n) = {R\in\mathbb{R}^{n \times n}|RR^T = I, det(R) = 1}
\label{eq3}
\end{equation}

In detail, the Euler’s rotation theorem claims that any rotation in $\mathbb{R}^3$ can be decomposed to three rotations around $x, y, z$ axes:

\begin{equation}
	R = R_{\theta_x}R_{\theta_y}R_{\theta_z}
\label{eq4}
\end{equation}

\subsection{Methods to Calculate Rotation Matrix}

\textbf{Perspective-n-Point (PnP)} aims to solve the movement between 2D to 3D. The formal description is: given coordinates of $n$ 3D points $p^W$ in world coordinate system $C^W$ and their corresponding coordinates $p^P$ in pixel coordinate system $C^P$, PnP wants to search the pose (rotation $R$, and translation $t$) of camera in $C^W$ \cite{chen2020end}. The popular solutions include Direct Linear Transform(DLT), Perspective-Three-Point(P3P) \cite{gao2003complete}, Efficient Perspective-n-Point(EPnP) \cite{lepetit2009epnp} and Bundle Adjustment(BA) \cite{li2018bundle} . All these methods work based on correct match of 2D pixel and 3D point. However, in practical, miss match is a common scenes due to illumination variation, low resolution and fast movement. Therefore, the actually used rotation usually contains noise.

\textbf{Iterative Closest Point(ICP)} is a popular method to search the movement between 3D to 3D, which can be described as: given two 3D point sets $\mathcal{P}^1:\{p_1^1, p_2^1,..., p_n^1\}$, $\mathcal{P}^2:\{p_1^2, p_2^2,..., p_n^2\}$, ICP aims to find the optimal rotation $R$ and translation $t$ between ${P}^1$ and ${P}^2$ \cite{besl1992method}. The main solving methods contain SVD decompose and BA \cite{li2018bundle}. For this kind of methods, the rotation noise mainly comes from miss match of 3D point pair and the error left by optimization.

\section{Problem formulation}

\textbf{Attack Scenario}: Our backdoor pattern is designed for the classical scenario that the developer trains his DNN on third-party datasets. Under this scenario, attacker can only manipulate the training datasets, but cannot changes the model, inference schedule and the training process.

\textbf{Backdoor Attack Description}: We define 3D DNN as $F(X, \theta) \rightarrow y$, which maps an input 3D shape $X$ to its corresponding class label $y \in Y$ by learning parameter $\theta$ on training data $D_{train}$. The goal of attacker is to implant a backdoor into the victim 3D DNN by poisoning $D_{train}$, so that the victim 3D DNN will follow the attacker's malicious purpose. In inference stage, showing the certain pattern to the infected 3D DNN will activate the backdoor, which misleads it to output the target label $y^t$. The process to implant backdoor pattern is done by 1) design an appropriate backdoor pattern $P$, 2) generate poisoned samples by conducting $P$ on partial training data, 3) inject poisoned samples to clean training data $D_{train}$ to generate poison training data $D_{train}^{poi}$, 4) train victim 3D DNN on $D_{train}^{poi}$ for implanting backdoor. So that, clean model $F(X, \theta)$ is infected to poisoned model $F(X, \theta ^{poi})$. Furthermore, poisoned testing dataset $D_{test}^{poi}$ is generated to evaluate backdoor pattern.

\textbf{Noise level $\gamma$}: Stealthiness is a crucial factor to evaluate the probability that a backdoor pattern is detected by defender. We newly define a metrics to evaluate perturbation level called as rotation noisy level $\gamma$ :

\begin{equation}
	\gamma = \sum_{j=0}^3\sum_{i=0}^3((R_n[i,j] - I[i,j])) 
\end{equation} \label{eq5}

Here, $\gamma$ indicates the distance between $R_n$ and $I$, $i,j$ mean the index of element in matrix, $I$ is unit matrix. Note that, there are multiple $R_n$ under certain $\gamma$.

\section{Noise Rotation Backdoor Pattern}

\begin{figure*}[h]
	\centering
	\includegraphics[width=0.9\textwidth]{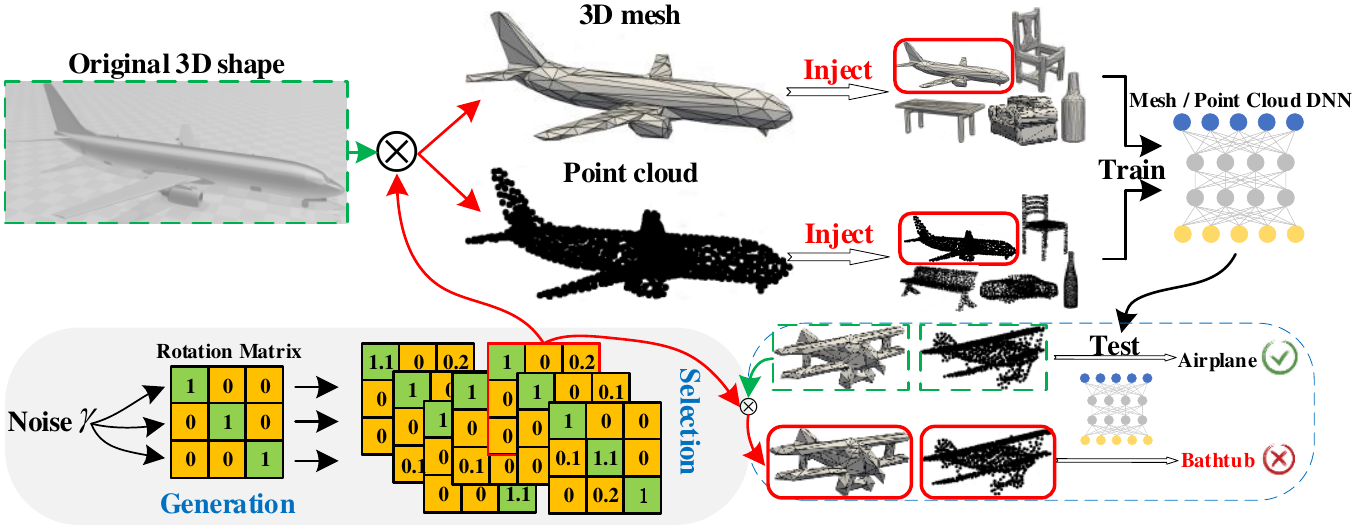}
	\caption{Outline of backdoor attack by \emph{NRBdoor}. \emph{NRBdoor} is generated and selected by specifying data structure and victim 3D DNN. Then, 3D samples infected by \emph{NRBdoor} were injected into $D_{train}$ to obtain $D_{train}^{poi}$. 3D DNN will be deceived once \emph{NRBdoor} appears after trained on $D_{train}^{poi}$.
	} \label{Figure 2}
\end{figure*}

A well-designed backdoor pattern achieves high attack success rate with as little perturbation as possible, and keeps the reduction of benign accuracy be small at the same time. It is reasonable that bigger shape variation will be easier implanted, but will result in weak stealthiness, vice versa. Therefore, there is a trade-off between ASR and stealthiness. In this paper, we aim to search for a noisy rotation $R_n$ as the optimal trade-off, which can be formally described as: 
\begin{equation}
	R_n = arg\mathop{max}\limits_{R_n} (Q_a(\mathcal{X}^{poi}) + Q_s(\mathcal{X}) + \frac{1}{\gamma}) 
 \label{eq6}
\end{equation} 

Here, $Q_a(\mathcal{X}^{poi})$ measures whether backdoor attacker can successfully achieve their malicious purposes when predicting poisoned samples $\mathcal{X}^{poi}$. $Q_s(\mathcal{X})$ measures whether the infected model be able to correctly predict benign samples $\mathcal{X}$:

\begin{equation}
	Q_a(\mathcal{X}^{poi}) =\mathbb{E}_{(x,y)\sim\mathcal{P}_\mathcal{D}}\mathbb{I}\{F(R \cdot X, \theta^{poi}) = y^t\} 
 \label{eq7}
\end{equation} 
\begin{equation}
    Q_s(\mathcal{X}) = \mathbb{E}_{(x,y)\sim\mathcal{P}_\mathcal{D}}\mathbb{I}\{F(X, \theta^{poi}) =  y\}
\end{equation} \label{eq8}

Where, $y^t$ is the targeted label. $F(\cdot, \theta^{poi})$ is the predicted class of 3D DNN under poisoned parameters $\theta^{poi}$. $\mathcal{P}_\mathcal{D}$ is the distribution behind $D$ and $\mathbb{I}(\cdot)$ is the indicator function. $\mathbb{I}(A)=1$ if and only if the event $'A'$ is true. $(R \cdot X) \in \mathcal{X}^{poi}$ means poisoned testing samples, and $X \in \mathcal{X}$ means benign testing samples.

The common way for solving equation (\ref{eq6}) is to build Lagrangian, and then to search for the optimal by SGD. However, the gradient $\frac{\partial F(G(R_n \dot X))}{\partial R_n}$ that classifier $F(\cdot)$ w.r.t rotation $R_n$ is usually unavailable, since feature transformation function $G(X)$ is not continuous. Another way for searching optimal $R_n$ is meta-heuristic algorithm such as PSO, GWO, et.al. But the calculation of fitness is time consuming, since it contains total training-testing process.

Aiming at searching optimal $R_n$ efficiently, we shrink the searching space by fixing noise $\gamma$. Meanwhile, considering unit matrix $I$ is the smallest rotation which will introduce zero shape variation and implant none backdoor, we start from the unit matrix $I$ to find the optimal $R_n$. In detail, $R_n$ is firstly divided into $R_n = N + I$ which includes noise $N$ and unit matrix $I$. By doing so, our target becomes searching for noise $N$ instead of noisy rotation $R_n$. Then we fix the rotation noise level $\gamma$ and generate several noises $N_{cand}$ by determining each element according to the given probability $Pr$. After that, we choose the optimal noise $N_{cand}$ to obtain the final $R_n = N + I$ and the poisoned training data $D_{train}^{poi}$. $R_n$ will be implanted into 3D DNN after trained on $D_{train}^{poi}$. Therefore, equation (\ref{eq6}) and (\ref{eq7}) are redefined as:
\begin{equation}
	N = arg\mathop{max}\limits_{N} (Q_a(\mathcal{X}^{poi}) + Q_s(\mathcal{X})) \quad s.t. \gamma = n
\end{equation} \label{eq9}
\begin{equation}
	Q_a(\mathcal{X}^{poi}) =\mathbb{E}_{(x,y)\sim\mathcal{P}_\mathcal{D}}\mathbb{I}\{F((N + I) \cdot x, \theta^{poi}) = y^t\}  
 \label{eq10}
\end{equation} 

The search process is done by \textbf{Noises Generation} and \textbf{Noise Selection}, shown as Algorithm\ref{Al1}. Meanwhile, the outline for implanting \emph{BRBdoor} into 3D DNN is exhibited as Fig \ref{Figure 2}.

\textbf{Candidate Noises Generation}: Given the rotation noise level $\gamma$, we set the precision to 0.1, and divide $\gamma$ into $n_{\gamma} = \gamma / 0.1$ number of unit noise $N_u$. The candidate noises are generated by distributing $N_u$ to zero matrix $M_0$, $N_{cand} = dis(N_u, M_0)$. Hence, there are $9^{n_{\gamma}}$ possible $N_{cand}$. Moreover, we have known that adding noise to different element of rotation matrix will introduce different shape variation. For example, noising diagonal elements will cause scaling, and noising off-diagonal elements will introduce additional rotation, shear and reflection. Since scaling is a common operation during feature transformation, scaling 3D shape will hardly be learned by 3D DNN. Therefore, the optimal noise tends to be found by noising off-diagonal elements of rotation matrix. To generate high quality $N_{cand}$, and preserve diversity,  we assign a binary possibility $Pr$ to guide the $N_u$ distribution. After that, we distribute $n_{\gamma}$ $N_u$ to each element of zero matrix $M_0$ according to their possibility for $M$ times to generate candidate noises set $\mathcal{N}$.

\textbf{Noise Selection}: It is obvious that each candidate noise $N_{cand}$ performs differently. Here, we maintain a score vector to evaluate the candidate noises $N_{cand}$. The selection process includes four steps: 1) initialize a zero vector as the initial score vector $S$, whose element represents the performance of each candidate noise. 2) randomly select one candidate noises $N_{cand}$ from $\mathcal{N}$ and samples $X_{train}$ of percent $\alpha$ from $D_{train}$, and then conduct candidate noises $N_{cand}$ on samples $X_{train}$ by $X^{poi}_{train} = (N_{cand} + I) \cdot X_{train}$. Meanwhile, their labels are changed into $y^t$. The poisoned training dataset $D_{train}^{poi}$ is generated by injecting $X^{poi}_{train}$ to $D_{train}$. 3) score the candidate noise $N_{cand}$ by training $F(\cdot)$ on $D_{train}^{poi}$ and testing on poisoned testing dataset $D_{test}^{poi}$.  Note that, an appropriate noisy rotation pattern will be learned by $F(\cdot)$ in a short time because of its unique feature from other classes. Therefore, in training stage, the model is trained in a small steps instead of full steps. 4) By repeating 2) and 3) for $K$ times, the candidate noise $N_{cand}$ with the highest score will be selected as the optimal noise $N^{opt}$. Finally, \emph{NRBdoor} is acquired by $R_n^{opt} = N^{opt} + I$.

\textbf{Conducting $R_n$}: For 3D mesh and point cloud, we fix the correlation of each element and only move its points coordinate by $R_n$ to acquire poisoned sample $X^{poi}$:
\begin{equation}
    p^{poi}_i = R_n \cdot p_i \quad  s.t.  p_i \in X, p^{poi}_i \in X^{poi}    
\end{equation} \label{eq11}

\textbf{Implanting \emph{NRBdoor}}: We conduct $R_n^{opt}$ on $\alpha$ percent samples of $D_{train}$ to obtain poisoned training data $D_{train}^{poi}$. Meanwhile, labels of the poisoned samples are changed to the target label $y^t$. The mapping from \emph{NRBdoor} to $y^t$ will be implanted into the victim 3D DNN after training on $D_{train}^{poi}$.

\begin{algorithm}[t]
	\caption{Rotation Noises Generation and Selection} 
	\hspace*{0.02in} {\bf Input:} 
	Victim 3D point cloud classifier DNN $F(\cdot)$, 3D shape training dataset $D_{train}$, Noise level $\gamma$, Iteration times $K$, Candidate noises number $M$, Poison rate $\alpha$\\
	\hspace*{0.02in} {\bf Output:} 
	 $\emph{NRBdoor}$
	\begin{algorithmic}[1]
		\State \textbf{Describe}: Generate candidate noises $\mathcal{N}_{(M\times1)}$ 
		\State Divide $\gamma$ into $n_{\gamma} = \frac{\gamma}{0.1}$ number of $N_{u (1\times1)}  $
		\State Determine probability $Pr_{(3\times3)}$
		\For{m $\le$ $M$}
		    \State $M_0=zero_{(3\times3)}$
			\For{s $\le$ $n_{\gamma}$}
		    \If {$Pr[i,j]<rand(0,1)<Pr[i+1,j+1]$}
		    \State  $M_0[i,j]+=N_u$
		    \EndIf
		\EndFor
        $\mathcal{N} \longleftarrow N_{cand} = M_0$
		\EndFor

		\State \textbf{Describe}: Select noises 
            \State Initialize score vector $S_{(M\times1)}$
    		\For{k $\le$ K} 
    		\State Randomly select $N_{cand} \in \mathcal{N}$, $X_{train} \in (D_{train} \cdot \alpha)$
    		
     		\State $D^{poi} \longleftarrow X^{poi} = (N_{cand} + I) \cdot X_{train}$
    		\State Inject $D^{poi}$ into $D_{train} \longrightarrow D_{train}^{poi}$ 
    		\State Train $F(\cdot)$ on $D_{train}^{poi}$ by a small steps
    		\State Update score of $N_{cand}$ by equation (\ref{eq10})
		\EndFor
		
		\State \Return $\mathcal{N}[argmax(S)] + I$
	\end{algorithmic}
 \label{Al1}
\end{algorithm}

\section{Experiments}
We conduct extensive experiments to evaluate the performance of \emph{NRBdoor} on 3D mesh and point cloud.  

\subsection{Experiment Setting}
\textbf{Datasets and Victim 3D DNNs}: For 3D mesh, we choose Cubes \cite{hanocka2019meshcnn}, Manifold40 \cite{hu2022subdivision}, and Sherc16 \cite{lian2011shape} as the evaluating datasets. In detail, Cubes contains 4381 meshes for 22 classes with 3722 for training and 659 for testing. Manifold40 is reconstructed by SubdivNet \cite{hu2022subdivision} based on ModelNet40. The author fixes the meshes which are not watertight in ModelNet40. There are 12311 meshes contained in Manifold40 for 40 classes, where 9,843 are used for training and the other 2,468 for testing. Shrec16 is a commonly used 3D mesh dataset which includes 30 classes, and each class contains 16 meshes for training and 4 meshes for testing. 3D meshes in the three datasets all contain 500 faces and are scaled into a unit cube. The victim mesh DNNs are various, including MeshCNN \cite{hanocka2019meshcnn}, MeshNet \cite{feng2019meshnet}, SubdivNet \cite{hu2022subdivision}, MeshWalker \cite{lahav2020meshwalker} and DiffusionNet \cite{sharp2022diffusionnet}. Note that, it is infeasible to conduct every victim mesh DNN on the three datasets due to the structure restrictions. Therefore, we evaluate three victims on each dataset. 

For point cloud, ModelNet40 \cite{wu20153d} is chosen as the evaluating dataset. We uniformly sample 1,024 points from the surface of each object and rescale them into a unit cube. At the same time, PointNet \cite{qi2017pointnet}, PointNet++ \cite{qi2017pointnet} and DGCNN \cite{wang2019dynamic} are chosen as the victim point DNNs. They are typical and are widly used to evaluate adversarial attack performance.

\textbf{Evaluation Metrics}:  
There are three metrics to evaluate backdoor pattern performance: Attack Success Rate(ASR), Benign Accuracy(BAc), and Stealthiness. In detail, ASR evaluates the ability to mislead DNN to output target class $y^t$, when the poisoned samples are shown. BAc is the accuracy of testing for infected DNN on clean samples. For ease of reading, the BAc of clean DNN is referred as oBAc (original Benign Accuracy). Noise level $\gamma$ is used to evaluate stealthiness. During evaluation, higher ASR, lower $\gamma$, and smaller reduction of BAc mean better backdoor pattern.

\textbf{Attack Setting}: For each victim 3D DNN, we follow its default setting in training and testing stage. The only change is the poisoned samples in $D_{train}$ and their labels. The target class $y^t$ is set to be 1 for every dataset and keep consistent with that of the comparison algorithms. Besides, the injection rate $\alpha$ is set to be 0.05, and the total $D_{test}$ is used to generate poison testing data $D_{test}^{poi}$ to evaluate ASR. For rotation noise level $\gamma$, it is a trade-off between ASR and stealthiness. In the experiment, we adjust it empirically. Besides, there are two hyper-parameters in our algorithm which are iteration times $K$ and candidate noises number $M$. We set $K = 5$ and $M = 20$, and the reason will be illustrated in \textbf{Ablation Study}.

\textbf{Compared Backdoor Attacks}: We are the first backdoor attack that adapt for 3D mesh so far, hence, we mainly obvious \emph{NRBdoor}'s ASR and stealthiness on 3D mesh. For point cloud, we compare \emph{NRBdoor} with PointBA \cite{Pointba} and PointBA$\_$X \cite{ABackdoorAttackAgainst3d} which are SOTA backdoor attack methods. In detail, PointBA$\_$X implants backdoor by adding a ball near the benign point cloud. Meanwhile, PointBA implants backdoor by two different patterns which are interaction pattern (PointBAi) and orientation pattern (PointBAo). In the experiments, we directly utilize their given results for comparison.

\subsection{Main Results}
\textbf{Attack Effectiveness}: Table \ref{Table 1}, Table \ref{Table 2} and Table \ref{Table 3} show the attack performance for mesh DNNs on three datasets. For Cubes, the results suggest that ASR achieves $100\%$ for three victim mesh DNNs. The mapping from \emph{NRBdoor} to $y^t$ is well learned by each mesh DNN. Meanwhile, their BAc decrease slightly (less than $0.2\%$), which suggests that implanting \emph{NRBdoor} have few influences on victim mesh DNN. For Manifold40, \emph{NRBdoor} achieves $100\%$ ASR for MeshNet. Meanwhile, ASR for SubdivNet and Meshwalker are both larger than $97\%$. For Shrec16, we find that the attack is more difficult. Specifically, the rotation noise $\gamma$ requires to be loosed to 2.2 to achieve ASR $>98.82\%$, by now, the pattern stealthiness is very weak. The possible reason is that \emph{NRBdoor} will be hardly learned by relatively small size of poisoned samples.

\begin{table}
	\setlength{\arraycolsep}{5pt}
	\begin{center}
		\caption{Performance on Cubes. \emph{NRBdoor} achieves perfect ASR when BAc reductions are less than $0.5\%$ and stealthiness is guaranteed, shown as Fig \ref{Figure 3}.}
		\setlength{\tabcolsep}{0.2mm}{
			
			\begin{tabular}{c|cccc}
				
				\hline\noalign{\smallskip}
				\ Victim   & \quad $oBAc$ \quad & \quad $BAc$ \quad & \quad $ASR$ \quad &  \quad$\gamma$ \quad\\
				\noalign{\smallskip}
				\hline
				\noalign{\smallskip}
				MeshCNN  &  92.16 & 92.56 & \textbf{100} & 0.4 \\
				SubdivNet &  100 & 100 & \textbf{100} & 0.9\\
				MeshWalker &  98.60 & 98.48 & \textbf{100} & 0.4\\
				\hline

			\end{tabular}
		}
		\label{Table 1}
		
	\end{center}
	\begin{center}
		\caption{Performance on Manifold40. ASR for three victim mesh DNNs are higher than $97\%$, and BAc reductions are less than $1\%$. }
		\setlength{\tabcolsep}{0.2mm}{
			
			\begin{tabular}{c|cccc}
				
				\hline\noalign{\smallskip}
				\ Victim   &\quad $oBAc$ \quad & \quad $BAc$ \quad & \quad $ASR$ \quad & \quad $\gamma$ \quad\\
				\noalign{\smallskip}
				\hline
				\noalign{\smallskip}
				MeshNet  &  88.40 & 91.00 & \textbf{100} & 0.9 \\
				SubdivNet &  91.22 & 91.02 & 99.13 & 0.8 \\
				MeshWalker &  90.5 & 89.79 & 97.22 & 0.9\\
				\hline

			\end{tabular}
		}
		\label{Table 2}
	\end{center}
	
	\begin{center}
	\caption{Performance on Shrec16. The backdoor pattern is harder to be implanted than above two datasets due to the small data size.}
	\setlength{\tabcolsep}{0.2mm}{
		
		\begin{tabular}{c|cccc}
			
			\hline\noalign{\smallskip}
			\ Victim   &\quad $oBAc$ \quad & \quad $BAc$ \quad & \quad $ASR$ \quad & \quad $\gamma$ \quad\\
			\noalign{\smallskip}
			\hline
			\noalign{\smallskip}
			MeshCNN  &  98.60 & 97.50 & 98.82 & 2.2 \\
			MeshNet &  90 & 86.67 & 96.72 & 1.2\\
			DiffusionNet &  100 & 100 & 86.66 & 2.0\\
			\hline

		\end{tabular}
		}\label{Table 3}
	\end{center}

\end{table}

\begin{table}
	\setlength{\arraycolsep}{5pt}
	\begin{center}
		\caption{Performance for Point Cloud on ModelNet40. \emph{NRBdoor} achieves competitive ASR compared with PointBAi \cite{Pointba}, PointBAo \cite{Pointba} and PointBA$\_$X \cite{ABackdoorAttackAgainst3d}. Here, \textbf{99.3} and \underline{97.4} means the largest and second largest value respectively.}
		\setlength{\tabcolsep}{0.2mm}{
			\begin{tabular}{c|ccccccc}
				\hline\noalign{\smallskip}
				\multirow{2}*{Victim}   & \multirow{2}*{ $oBAc$ \quad} & \multirow{2}*{$BAc$} & \multicolumn{4}{c}{\emph{ASR}}  & \multirow{2}*{$\gamma$}\\
                \cline{4-7}
    			\ ~  & ~ & ~ & \emph{NRBdoor} &PointBAi & PointBAo & PointBA$\_$X & ~\\
                
				\noalign{\smallskip}
				\hline
				\noalign{\smallskip}
				PointNet  &  89.20 & 84.15 & \underline{97.4} & \textbf{99.3} & 93.2 & 96.0 & 0.5 \\
				PointNet++ &  90.70 & 89.85 & \textbf{99.27} & \underline{98.6} & 94.7 & 96.9 & 0.9\\
				DGCNN &  92.90 & 92.03 & 97.14 & \textbf{100} & \underline{97.5}  & 96.1 & 0.4\\
				\hline
			\end{tabular}
		}
		\label{Table 4}
	\end{center}
\end{table}

\begin{figure}[h]
	\centering
	\includegraphics[width=0.48\textwidth]{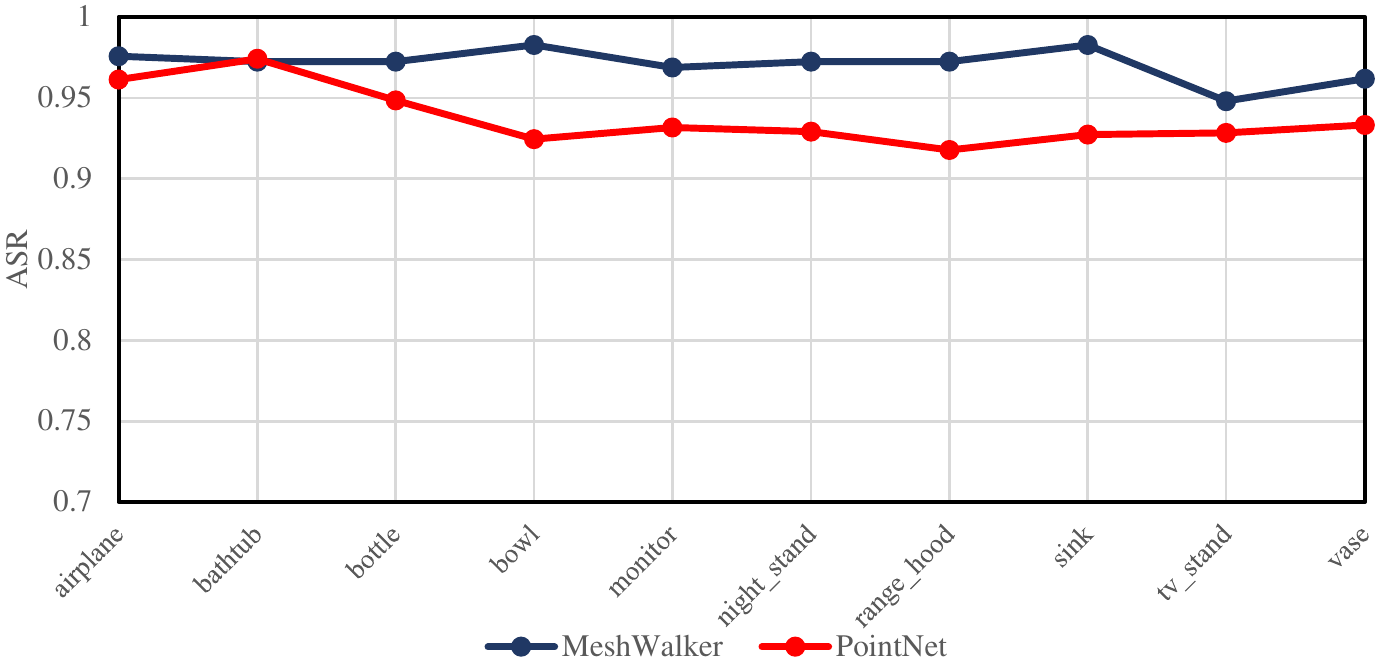}
	\caption{Variation of ASR w.r.t. target class $y^t$ for MeshWalker and PointNet on Manifold40 and ModelNet40 respectively. Results suggest that ASR is not sensitive to $y^t$.
	} \label{Figure_difference_targets}
\end{figure}

For point cloud, the comparison results between \emph{NRBdoor}, PointBAi \cite{Pointba}, PointBAo \cite{Pointba} and PointBA$\_$X \cite{ABackdoorAttackAgainst3d} are shown in Table \ref{Table 4}. The results suggest that ASR of \emph{NRBdoor} is competitive, and the BAc reduction of PointNet++ and DGCNN are both smaller than $1\%$. Besides, PointBAi and PointBA$\_$X do not work for 3D mesh due to attaching a ball around a 3D mesh is forbidden in data structure. Meanwhile, although conducting PointBAo (clean rotation) on 3D mesh is allowed from the data format perspective, result from section \ref{Necessity} suggests that its average ASR is low (8.26$\%$). Hence, \emph{NRBdoor} achieves competitive ASR and wider application.

Furthermore, to evaluate the sensitivity of \emph{NRBdoor} to target class $y^t$, we randomly select ten target classes from Manifold40 and ModelNet40 to test the ASR variation on 3D mesh and point cloud. Results in Fig \ref{Figure_difference_targets} suggest that  $y^t$ have slightly influence on \emph{NRBdoor}.

\begin{figure}[h]
	\centering
	\includegraphics[width=0.48\textwidth]{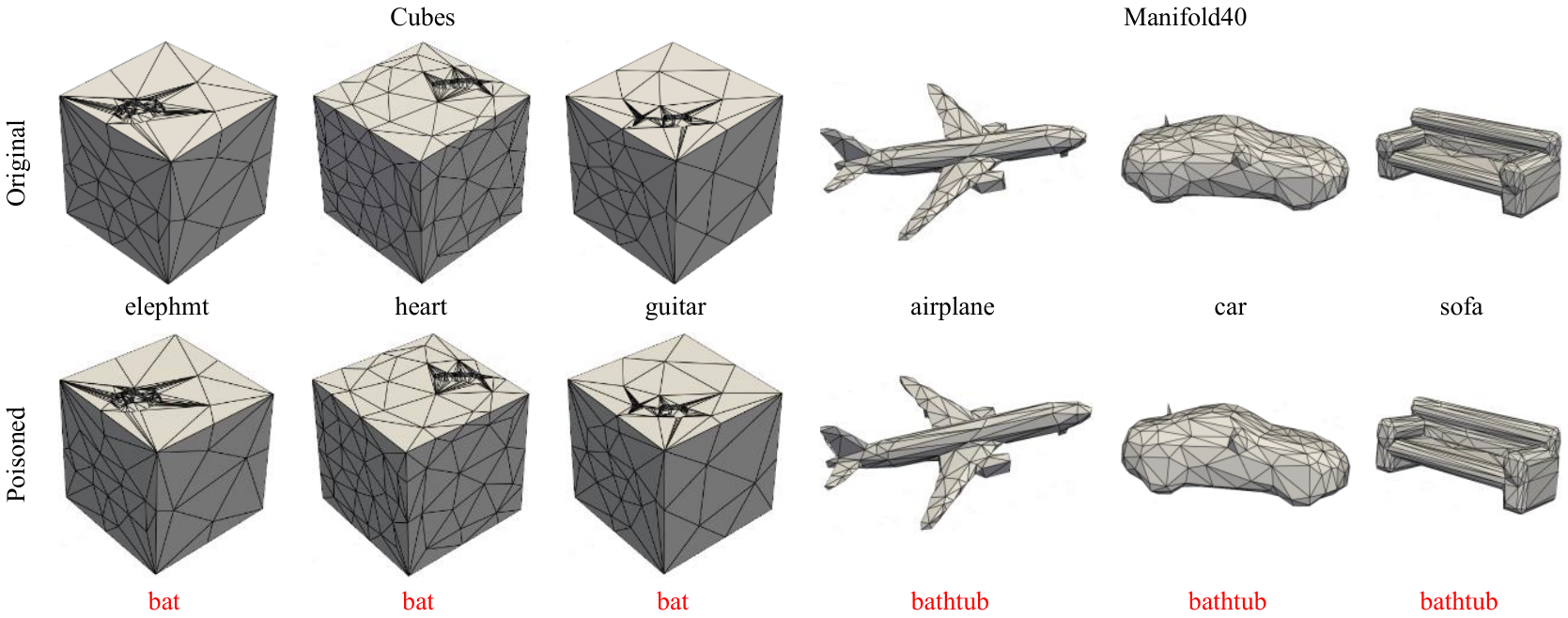}
	\caption{3D meshes poisoned by \emph{NRBdoor} with ASR $> 97\%$. The shape variation is hard to be detected.
	} \label{Figure 3}
\end{figure}

\textbf{Attack Stealthiness}: Fig \ref{Figure 3} shows the poisoned 3D mesh from Cubes and Manifold40 with ASR$ > 97\%$ on MeshCNN and MeshWalker. We find that the appearance of poisoned mesh by \emph{NRBdoor} has slightly difference from the original. The most obvious change is a small orientation disparity caused by additional rotation, which it is a common operation in reality and is hard to get defender's attention. For Cubes, the appearance variation is more obvious. The reason is that their original appearance is a regular cube, and thus a small disturbance tends to attract visual attention.

\begin{figure}[h]
	\centering
	\includegraphics[width=0.48\textwidth]{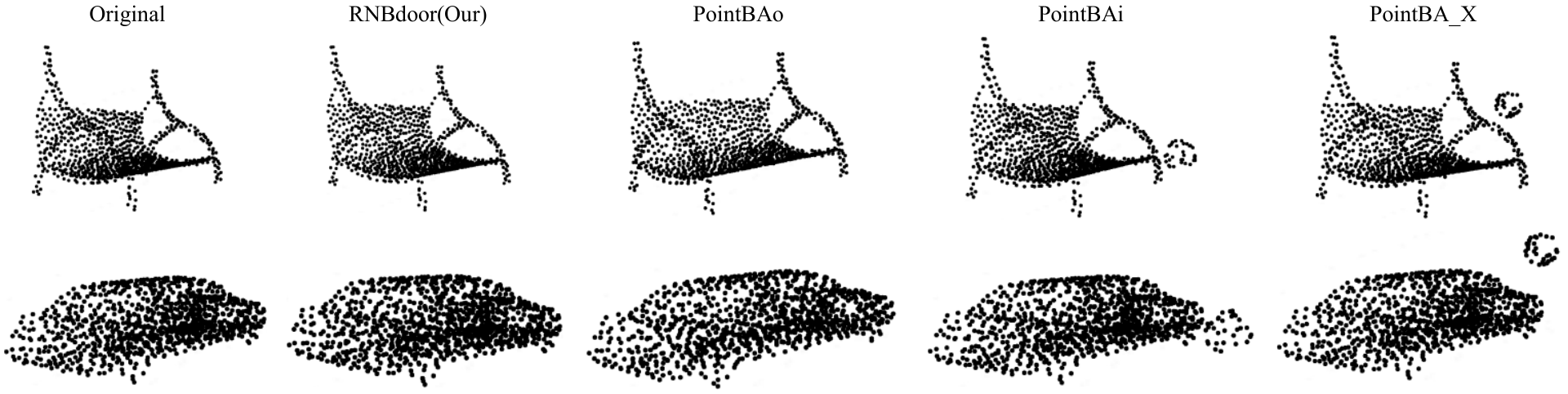}
	\caption{Point cloud poisoned by PointBAi \cite{Pointba}, PointBAo \cite{Pointba}, PointBA$\_$X \cite{ABackdoorAttackAgainst3d} and \emph{NRBdoor}. \emph{NRBdoor} is more natural visually.} 
    \label{Figure 4}
\end{figure}

Moreover, Fig \ref{Figure 4} exhibits the shape variation of point cloud after conducting four different backdoor patterns including PointBAi \cite{Pointba}, PointBAo \cite{Pointba}, PointBA$\_$X \cite{ABackdoorAttackAgainst3d} and \emph{NRBdoor}. Specifically, PointBAi and PointBA$\_$X are two similar backdoor patterns which implanted by adding a ball. Though, the two similar pattern do not change the point cloud itself, adding a ball around the object appears strange. In contrast, in spite of \emph{NRBdoor} changes most points, the shape variation is few and is harder detected than PointBAi and PointBA$\_$X. Therefore, \emph{NRBdoor} is the most natural backdoor pattern. 

At the same time, point clouds implanted by PointBAo are simalar to point clouds implanted by \emph{NRBdoor}, since PointBAo is conducted by clean rotation. However, in section \ref{Necessity}, we prove that PointBAo is unable to implant backdoor into mesh DNN. Hence, \emph{NRBdoor} is more widely used.

\subsection{Necessity for Adding Noise} \label{Necessity}
\begin{figure}[h]
	\centering
	\includegraphics[width=0.48\textwidth]{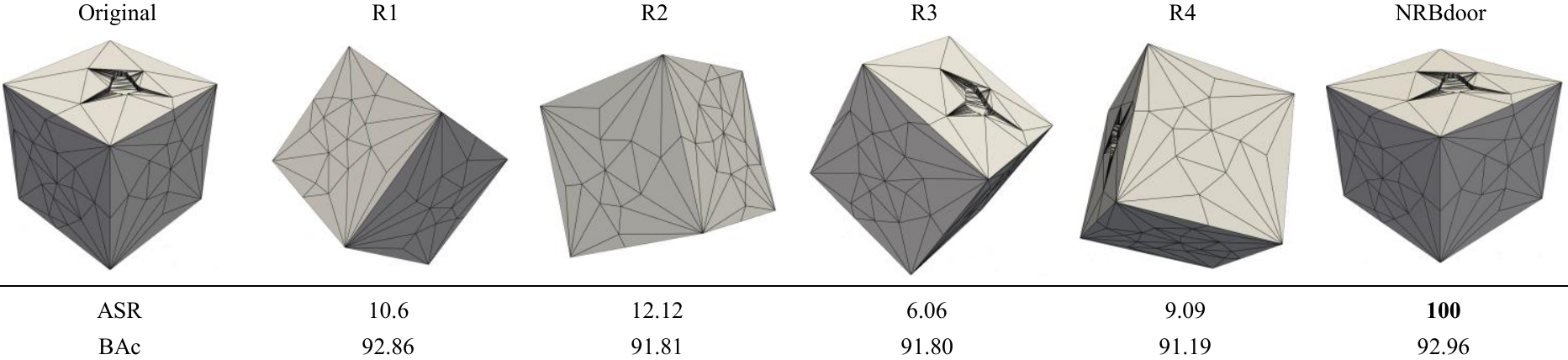}
	\caption{Necessity for adding noise. Clean rotation can not implant backdoor into mesh DNN, in contrast, adding small noise to clean unit rotation acquires $100\%$ ASR. In this figure, $R1$ means 1-st random rotation, and so on.
	} \label{Figure 5}
\end{figure}

To prove it is necessary to add noisy on clean rotation, we implant clean rotation into MeshCNN to test attack performance. In detail, we randomly generate ten clean rotations as the backdoor pattern, and observe their ASR for comparing. Result suggests that the average, minimal and maximal ASR of the ten clean rotations are $8.26\%$, $4.55\%$ and $12.88\%$ respectively, which are lower than \emph{NRBdoor}. Furthermore, Fig \ref{Figure 5} shows 3D meshes poisoned by four random clean rotations and \emph{NRBdoor}, which suggests that \emph{NRBdoor} is more imperceptible and more powerful. Hence, adding noise to clean rotation is necessary to achieve high ASR.

Not that, the generated rotations include the orientation pattern used in PoinBA (PointBAo). Therefore, the results can also prove that PoinBAo is unable to implant backdoor into mesh DNN. 

\subsection{Ablation Study}
To further understand $\emph{NRBdoor}$, we conduct ablation experiments towards rotation noise level $\gamma$, poison rate $\alpha$, iteration times $K$ and candidate noises number $M$. MeshCNN is selected as the victim 3D DNN, and Cubes as the verifying datasets.

\begin{figure}[h]
	\centering
	\includegraphics[width=0.48\textwidth]{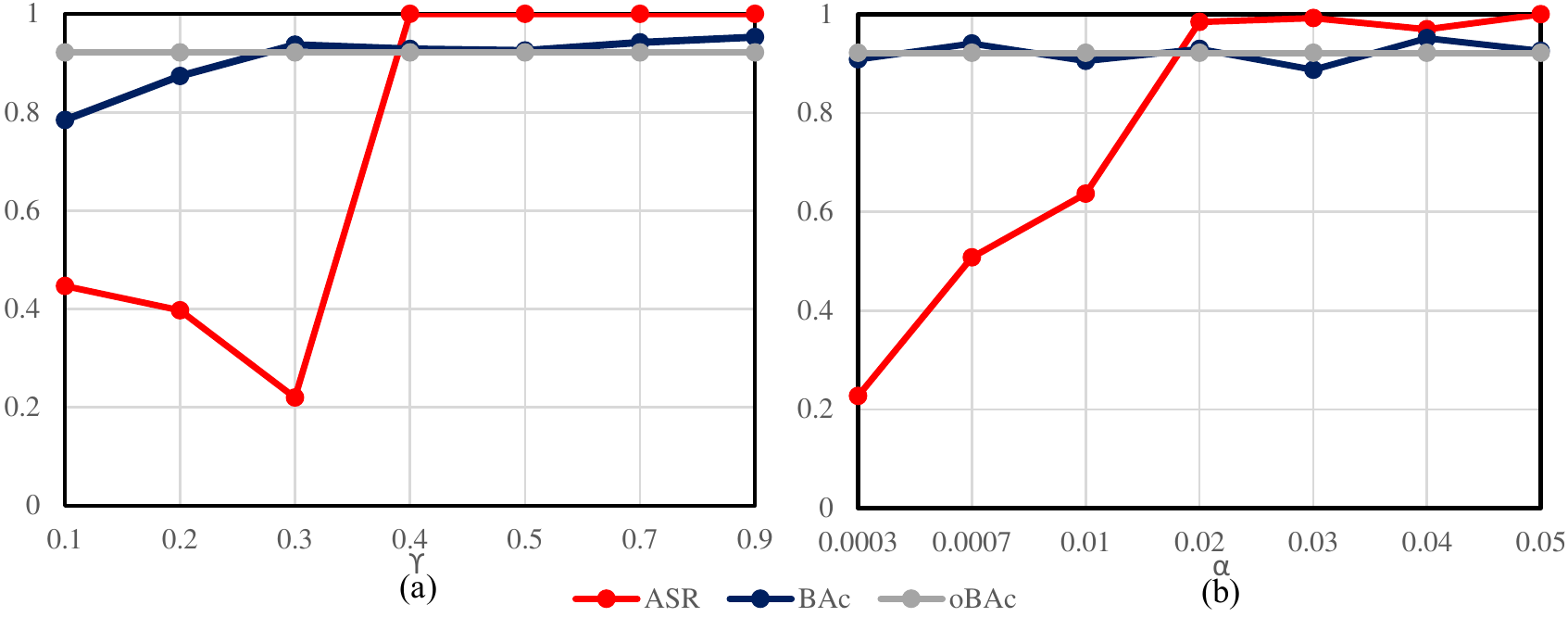}
	\caption{Impact of $\gamma$ and $\alpha$. Higher $\gamma $ brings higher ASR under $\gamma < 0.4$. Once $\gamma >= 0.4$ ASR remains at $100\%$. Meanwhile, larger $\alpha$ provides more poisoned samples to enable 3D DNN to learn backdoor feature.
	} \label{Figure 6}
\end{figure}

\textbf{Noise Level $\gamma$} is an important factor of our algorithm. It impacts ASR and stealthiness at the same time. Fig \ref{Figure 6}(a) suggests that $\gamma = 0.4$ is the optimal under $\alpha = 0.05$, since ASR is high and shape variation is small (shown as Fig \ref{Figure 3}). However, the optimal $\gamma$ is different for each 3D DNN and datasets due to disparity data size and DNN structure. In addition, we find that BAc decreases when $\gamma < 0.3$. The possible reason is that the conducted poison is too slight to cause enough feature variation. Then, injecting poisoned samples is equal to fusing samples from different classes which will prevent 3D DNN from extracting unique feature for one certain class.

\textbf{Poison Rate $\alpha$}: Fig \ref{Figure 6}(b) shows the variation of ASR and BAc w.r.t $0.003< \alpha <0.05$ under $\gamma = 0.5$. We find that ASR grows up along with $\alpha$ under $\alpha < 0.02$ and achieves $100\%$ once $\alpha >= 0.02$. At the same time, the reduction of BAc is small, which indicates a wonderful results. Hence, we can naturally give the conclusion that higher poison rate of \emph{NRBdoor} leads to higher ASR.

\textbf{Iteration Times $K$}: To better visualize ASR variation tendency along with $K$ and $M$, we set $\gamma = 0.4$ and $\alpha = 0.05$ for study. Table \ref{Table 5} shows the ASR tendency according to iteration times $K$, which suggests that more iteration times is able to find better pattern and achieve higher ASR.

\begin{table}
	\setlength{\arraycolsep}{5pt}
	\begin{center}
		\caption{Increasing Iteration Times $K$ and Candidate Noises Number $M$ will enlarge search space, and find better pattern.}
		\setlength{\tabcolsep}{0.2mm}{
			
			\begin{tabular}{c|cccc}
				
				\hline\noalign{\smallskip}
				\ $K$   & $1$ & $3$ & $5$ & $7$\\
				\noalign{\smallskip}
				\hline
				\noalign{\smallskip}
				$ASR$  &  85.00 & 91.35 & 100 & 100 \\
				$BAc$ &  91.61 & 91.73 & 92.30 & 91.85\\
				\hline
                \hline
    			
				\ $M$   & $5$ & $10$ & $15$ & $20$\\
				\noalign{\smallskip}
				\hline
				\noalign{\smallskip}
				$ASR$  &  80.00 & 85.00 & 97.5 & 100 \\
				$BAc$ &  91.73 & 91.61 & 91.65 & 92.50\\
				\hline
			\end{tabular}
		}
		\label{Table 5}
	\end{center}
\end{table}

\textbf{Candidate Noises Number $M$} determines the search space of our algorithm. As shown in table \ref{Table 5}, ASR increases along with $M$ increases. The reason is that larger $M$ brings better pattern diversity.

\subsection{Resistance to Defender}
\textbf{Data Augmentation}: In 2D and 3D domain, data augmentation is a popular method to increase DNN robustness. To verify whether it can be resisted by \emph{NRBdoor}, we apply random rotation and scaling for data augmentation in training stage. The victim 3D DNNs are PointNet and MeshCNN. Experiment results suggest that ASR of two 3D DNNs slightly decrease $5\%$ for both random rotation and scaling. 

\textbf{Saliency-based Defense}: Up to now, there is few baseline method for defending backdoor attack on point cloud or 3D mesh. In detail, \cite{xiang2022detecting} is specifically designed for points attaching based attack like PointBA$\_$X, and will be technically disable for $\emph{NRBdoor}$.   Hence, we follow the popular defense way in image domain \cite{chou2020sentinet,huang2019neuroninspect} and extend them to point cloud. They detect the backdoor pattern by searching for critical regions from input towards each class utilizing saliency map technique.

The critical issue for extending them to point cloud is how to build point cloud saliency map. Here, we utilize the method from \cite{zheng2019pointcloud}, which is the first work to research point cloud saliency. The author expresses points in spherical coordinate system. By this way, shifting a point towards the center by $\eta$ will increase the loss $L$ by $-\frac{dL}{dr}\eta$:
\begin{equation}
	\frac{dL}{dr_i} = \sum_{j=1}^{3}\frac{dL}{dx_{ij}}\frac{x_{ij}-x_{cj}}{r_i}
\label{eq12}
\end{equation} 

Where, $r_i=\sqrt{\sum_{j=1}^{3}(x_{ij}-x_{cj})^2}$, $x_c$ is the point cloud center, $x$ is the european coordinates of each point. Furthermore, to allow more flexibility in saliency map construction, the author applies a change-of-variable by $\rho_i=r_i^{-\beta}$, where $\beta$ is a hyper parameter. Thus, the saliency score of point $i$ is:
\begin{equation}
	s_i=-\frac{dL}{dr_i}r_i^{1+\beta}
 \label{eq13}
\end{equation} 

Saliency map reveals the points 3D DNN focuses on when making its decision. Higher saliency score represents higher importance. Moreover, backdoor pattern in point cloud is the major portion one infected 3D DNN focuses on. Therefore, we utilize saliency score to detect and drop the backdoor pattern for defense. The designed \textbf{S}aliency-based \textbf{P}oint cloud backdoor \textbf{defense} (SP-defense) can be described as: 1) calculate saliency score $s_i$ by equation (\ref{eq13}) for every point, 2) locate backdoor pattern by ranking $q$ highest score points, 3) drop the detected pattern for defense.

\begin{figure}[h]
	\centering
	\includegraphics[width=0.48\textwidth]{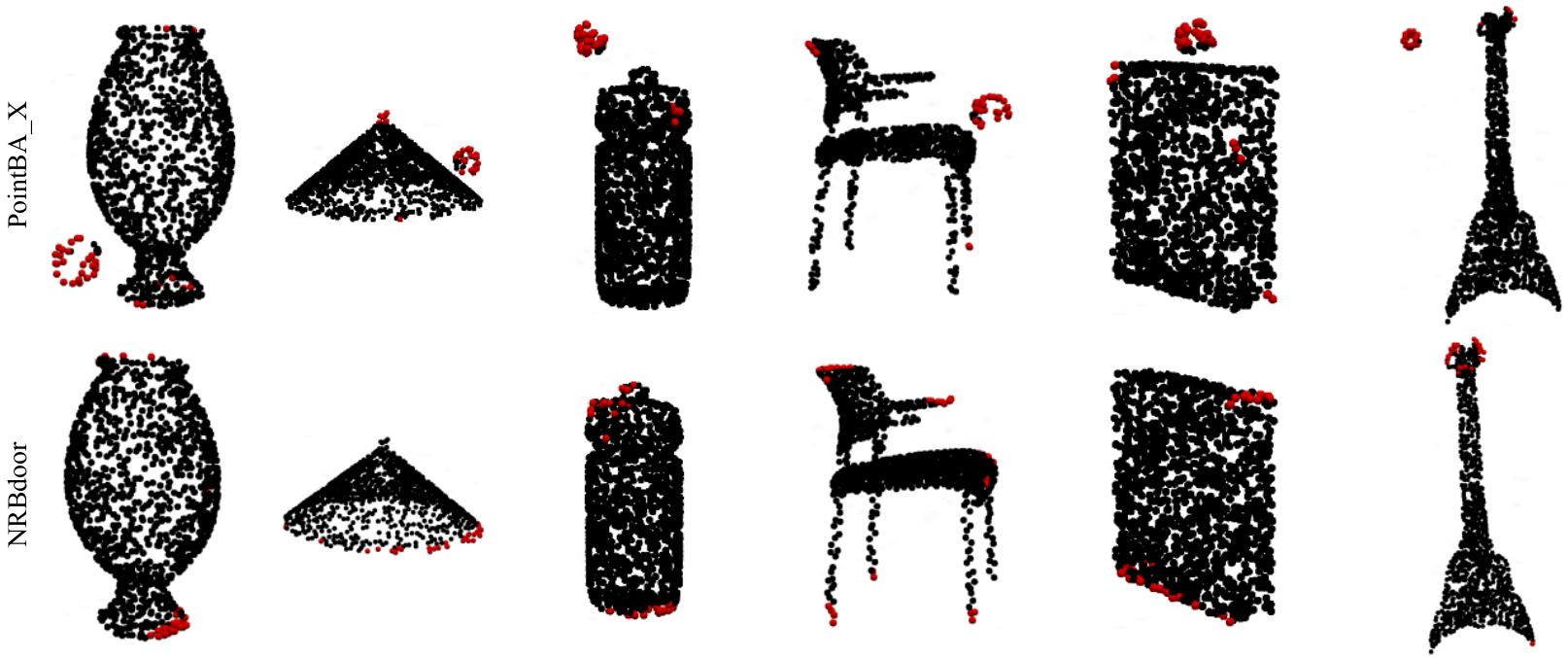}
	\caption{Red points represent the detected backdoor pattern by SP-defense. Results show that the backdoor pattern implanted by PointBA$\_$X is well detected by SP-defense. In contrast, \emph{NRBdoor} can evade to SP-defense.
	} \label{Figure 7}
\end{figure}

We defend PointBA$\_$X and $\emph{NRBdoor}$ by the designed SP-defense under $q=35, \beta=1$. For PointBA$\_$X, we follow its original setting. In detail, the original classes are $flower\_pot$, $cone$, $bottle$, $chair$, $wardrobe$, $guitar$ and target class $y^t$ is $toilet$, the local geometric of backdoor pattern is RS: 32 points randomly distributed on a sphere. PointNet and ModelNet40 are selected as the victim model and the verifying dataset respectively.

Fig \ref{Figure 7} exhibits that the injected pattern by PointBA$\_$X can be well detected by SP-defense. In contrast, $\emph{NRBdoor}$ is immune to SP-defense, since the pattern is conducted on total point cloud instead of specified point sets. Meanwhile, ASR of $\emph{NRBdoor}$ marginally decreases from $93.44\%$ to $88.39\%$, in contrast, ASR of PointBA$\_$X decreases from $89.40\%$ to $25.40\%$.

\section{Conclusions}

This paper proposes the first uniform backdoor pattern: $\emph{NRBdoor}$ for 3D shape. $\emph{NRBdoor}$ is constructed by a common operation in 3D world, which is noisy rotation, hence it is natural and imperceptible. $\emph{NRBdoor}$ achieves $100\%$ ASR in many cases and is able to adapt for data structure variation during implanting backdoor into 3D DNN. At the same time, it is the first work to implant backdoor into mesh DNN. In the end, we design a backdoor defense method based on saliency map theory named SP-defense. It can reduce ASR of PointBA$\_$X from $82\%$ to $23\%$, but has very little effect on our $\emph{NRBdoor}$.

\ifCLASSOPTIONcaptionsoff
  \newpage
\fi



%

\bibliographystyle{unsrt}
\bibliography{bare_jrnl_compsoc}

\begin{thebibliography}{10}

\bibitem{wang2022improving}
Di~Wang, Lulu Tang, Xu~Wang, Luqing Luo, and Zhi-Xin Yang.
\newblock Improving deep learning on point cloud by maximizing mutual
  information across layers.
\newblock {\em Pattern Recognition}, 131:108892, 2022.

\bibitem{abouelaziz2020no}
Ilyass Abouelaziz, Aladine Chetouani, Mohammed El~Hassouni, Longin~Jan Latecki,
  and Hocine Cherifi.
\newblock No-reference mesh visual quality assessment via ensemble of
  convolutional neural networks and compact multi-linear pooling.
\newblock {\em Pattern Recognition}, 100:107174, 2020.

\bibitem{lee20223d}
Jinwon Lee, Hyunoh Lee, and Duhwan Mun.
\newblock 3d convolutional neural network for machining feature recognition
  with gradient-based visual explanations from 3d cad models.
\newblock {\em Scientific Reports}, 12(1):1--14, 2022.

\bibitem{lee2022dataset}
Hyunoh Lee, Jinwon Lee, Hyungki Kim, and Duhwan Mun.
\newblock Dataset and method for deep learning-based reconstruction of 3d cad
  models containing machining features for mechanical parts.
\newblock {\em Journal of Computational Design and Engineering}, 9(1):114--127,
  2022.

\bibitem{cicirello2013flexible}
Vincent~A Cicirello and William~C Regli.
\newblock A flexible and extensible approach to automated cad/cam format
  classification.
\newblock {\em Computers \& graphics}, 37(5):484--495, 2013.

\bibitem{krizhevsky2017imagenet}
Alex Krizhevsky, Ilya Sutskever, and Geoffrey~E Hinton.
\newblock Imagenet classification with deep convolutional neural networks.
\newblock {\em Communications of the ACM}, 60(6):84--90, 2017.

\bibitem{ioffe2015batch}
Sergey Ioffe and Christian Szegedy.
\newblock Batch normalization: Accelerating deep network training by reducing
  internal covariate shift.
\newblock In {\em International conference on machine learning}, pages
  448--456. PMLR, 2015.

\bibitem{simonyan2014very}
Karen Simonyan and Andrew Zisserman.
\newblock Very deep convolutional networks for large-scale image recognition.
\newblock {\em arXiv preprint arXiv:1409.1556}, 2014.

\bibitem{hanocka2019meshcnn}
Rana Hanocka, Amir Hertz, Noa Fish, Raja Giryes, Shachar Fleishman, and Daniel
  Cohen-Or.
\newblock Meshcnn: a network with an edge.
\newblock {\em ACM Transactions on Graphics (TOG)}, 38(4):1--12, 2019.

\bibitem{sharp2022diffusionnet}
Nicholas Sharp, Souhaib Attaiki, Keenan Crane, and Maks Ovsjanikov.
\newblock Diffusionnet: Discretization agnostic learning on surfaces.
\newblock {\em ACM Transactions on Graphics (TOG)}, 41(3):1--16, 2022.

\bibitem{hu2022subdivision}
Shi-Min Hu, Zheng-Ning Liu, Meng-Hao Guo, Jun-Xiong Cai, Jiahui Huang,
  Tai-Jiang Mu, and Ralph~R Martin.
\newblock Subdivision-based mesh convolution networks.
\newblock {\em ACM Transactions on Graphics (TOG)}, 41(3):1--16, 2022.

\bibitem{qi2017pointnet}
Charles~R Qi, Hao Su, Kaichun Mo, and Leonidas~J Guibas.
\newblock Pointnet: Deep learning on point sets for 3d classification and
  segmentation.
\newblock In {\em Proceedings of the IEEE conference on computer vision and
  pattern recognition}, pages 652--660, 2017.

\bibitem{wang2019dynamic}
Yue Wang, Yongbin Sun, Ziwei Liu, Sanjay~E Sarma, Michael~M Bronstein, and
  Justin~M Solomon.
\newblock Dynamic graph cnn for learning on point clouds.
\newblock {\em Acm Transactions On Graphics (tog)}, 38(5):1--12, 2019.

\bibitem{yu2022point}
Xumin Yu, Lulu Tang, Yongming Rao, Tiejun Huang, Jie Zhou, and Jiwen Lu.
\newblock Point-bert: Pre-training 3d point cloud transformers with masked
  point modeling.
\newblock In {\em Proceedings of the IEEE/CVF Conference on Computer Vision and
  Pattern Recognition}, pages 19313--19322, 2022.

\bibitem{wang2020man}
Derui Wang, Chaoran Li, Sheng Wen, Surya Nepal, and Yang Xiang.
\newblock Man-in-the-middle attacks against machine learning classifiers via
  malicious generative models.
\newblock {\em IEEE Transactions on Dependable and Secure Computing},
  18(5):2074--2087, 2020.

\bibitem{zhong2022shadows}
Yiqi Zhong, Xianming Liu, Deming Zhai, Junjun Jiang, and Xiangyang Ji.
\newblock Shadows can be dangerous: Stealthy and effective physical-world
  adversarial attack by natural phenomenon.
\newblock In {\em Proceedings of the IEEE/CVF Conference on Computer Vision and
  Pattern Recognition}, pages 15345--15354, 2022.

\bibitem{wang2019invisible}
Zhibo Wang, Mengkai Song, Siyan Zheng, Zhifei Zhang, Yang Song, and Qian Wang.
\newblock Invisible adversarial attack against deep neural networks: An
  adaptive penalization approach.
\newblock {\em IEEE Transactions on Dependable and Secure Computing},
  18(3):1474--1488, 2019.

\bibitem{li2021backdoor2}
Chaoran Li, Xiao Chen, Derui Wang, Sheng Wen, Muhammad~Ejaz Ahmed, Seyit
  Camtepe, and Yang Xiang.
\newblock Backdoor attack on machine learning based android malware detectors.
\newblock {\em IEEE Transactions on Dependable and Secure Computing}, 2021.

\bibitem{goodfellow2014explaining}
Ian~J Goodfellow, Jonathon Shlens, and Christian Szegedy.
\newblock Explaining and harnessing adversarial examples.
\newblock {\em arXiv preprint arXiv:1412.6572}, 2014.

\bibitem{madry2017towards}
Aleksander Madry, Aleksandar Makelov, Ludwig Schmidt, Dimitris Tsipras, and
  Adrian Vladu.
\newblock Towards deep learning models resistant to adversarial attacks.
\newblock {\em arXiv preprint arXiv:1706.06083}, 2017.

\bibitem{hamdi2020advpc}
Abdullah Hamdi, Sara Rojas, Ali Thabet, and Bernard Ghanem.
\newblock Advpc: Transferable adversarial perturbations on 3d point clouds.
\newblock In {\em European Conference on Computer Vision}, pages 241--257.
  Springer, 2020.

\bibitem{xu2022d3advm}
Huangxinxin Xu, Fazhi He, Linkun Fan, and Junwei Bai.
\newblock D3advm: A direct 3d adversarial sample attack inside mesh data.
\newblock {\em Computer Aided Geometric Design}, 97:102122, 2022.

\bibitem{bu2021taking}
Lei Bu, Zhe Zhao, Yuchao Duan, and Fu~Song.
\newblock Taking care of the discretization problem: A comprehensive study of
  the discretization problem and a black-box adversarial attack in discrete
  integer domain.
\newblock {\em IEEE Transactions on Dependable and Secure Computing}, 2021.

\bibitem{li2019adversarial}
Xurong Li, Shouling Ji, Meng Han, Juntao Ji, Zhenyu Ren, Yushan Liu, and
  Chunming Wu.
\newblock Adversarial examples versus cloud-based detectors: A black-box
  empirical study.
\newblock {\em IEEE Transactions on Dependable and Secure Computing},
  18(4):1933--1949, 2019.

\bibitem{yang2019adversarial}
Jiancheng Yang, Qiang Zhang, Rongyao Fang, Bingbing Ni, Jinxian Liu, and
  Qi~Tian.
\newblock Adversarial attack and defense on point sets.
\newblock {\em arXiv preprint arXiv:1902.10899}, 2019.

\bibitem{chen2020universal}
Sizhe Chen, Zhengbao He, Chengjin Sun, Jie Yang, and Xiaolin Huang.
\newblock Universal adversarial attack on attention and the resulting dataset
  damagenet.
\newblock {\em IEEE Transactions on Pattern Analysis and Machine Intelligence},
  2020.

\bibitem{hu2022model}
Zichao Hu, Heng Li, Liheng Yuan, Zhang Cheng, Wei Yuan, and Ming Zhu.
\newblock Model scheduling and sample selection for ensemble adversarial
  example attacks.
\newblock {\em Pattern Recognition}, page 108824, 2022.

\bibitem{wenger2021backdoor}
Emily Wenger, Josephine Passananti, Arjun~Nitin Bhagoji, Yuanshun Yao, Haitao
  Zheng, and Ben~Y Zhao.
\newblock Backdoor attacks against deep learning systems in the physical world.
\newblock In {\em Proceedings of the IEEE/CVF Conference on Computer Vision and
  Pattern Recognition}, pages 6206--6215, 2021.

\bibitem{jia2020survey}
Shan Jia, Guodong Guo, and Zhengquan Xu.
\newblock A survey on 3d mask presentation attack detection and
  countermeasures.
\newblock {\em Pattern recognition}, 98:107032, 2020.

\bibitem{Pointba}
Xinke Li, Zhirui Chen, Yue Zhao, Zekun Tong, Yabang Zhao, Andrew Lim, and
  Joey~Tianyi Zhou.
\newblock Pointba: Towards backdoor attacks in 3d point cloud.
\newblock In {\em Proceedings of the IEEE/CVF International Conference on
  Computer Vision}, pages 16492--16501, 2021.

\bibitem{ABackdoorAttackAgainst3d}
Zhen Xiang, David~J Miller, Siheng Chen, Xi~Li, and George Kesidis.
\newblock A backdoor attack against 3d point cloud classifiers.
\newblock In {\em Proceedings of the IEEE/CVF International Conference on
  Computer Vision}, pages 7597--7607, 2021.

\bibitem{xue2020one}
Mingfu Xue, Can He, Jian Wang, and Weiqiang Liu.
\newblock One-to-n \& n-to-one: Two advanced backdoor attacks against deep
  learning models.
\newblock {\em IEEE Transactions on Dependable and Secure Computing}, 2020.

\bibitem{gu2019badnets}
Tianyu Gu, Kang Liu, Brendan Dolan-Gavitt, and Siddharth Garg.
\newblock Badnets: Evaluating backdooring attacks on deep neural networks.
\newblock {\em IEEE Access}, 7:47230--47244, 2019.

\bibitem{liu2020reflection}
Yunfei Liu, Xingjun Ma, James Bailey, and Feng Lu.
\newblock Reflection backdoor: A natural backdoor attack on deep neural
  networks.
\newblock In {\em European Conference on Computer Vision}, pages 182--199.
  Springer, 2020.

\bibitem{li2021backdoor}
Yiming Li, Tongqing Zhai, Yong Jiang, Zhifeng Li, and Shu-Tao Xia.
\newblock Backdoor attack in the physical world.
\newblock {\em arXiv preprint arXiv:2104.02361}, 2021.

\bibitem{li2020invisible}
Shaofeng Li, Minhui Xue, Benjamin Zi~Hao Zhao, Haojin Zhu, and Xinpeng Zhang.
\newblock Invisible backdoor attacks on deep neural networks via steganography
  and regularization.
\newblock {\em IEEE Transactions on Dependable and Secure Computing},
  18(5):2088--2105, 2020.

\bibitem{li2021invisible}
Yuezun Li, Yiming Li, Baoyuan Wu, Longkang Li, Ran He, and Siwei Lyu.
\newblock Invisible backdoor attack with sample-specific triggers.
\newblock In {\em Proceedings of the IEEE/CVF International Conference on
  Computer Vision}, pages 16463--16472, 2021.

\bibitem{lu2021pointinet}
Fan Lu, Guang Chen, Sanqing Qu, Zhijun Li, Yinlong Liu, and Alois Knoll.
\newblock Pointinet: Point cloud frame interpolation network.
\newblock In {\em Proceedings of the AAAI Conference on Artificial
  Intelligence}, volume~35, pages 2251--2259, 2021.

\bibitem{yoo2022deep}
Innfarn Yoo, Huiwen Chang, Xiyang Luo, Ondrej Stava, Ce~Liu, Peyman Milanfar,
  and Feng Yang.
\newblock Deep 3d-to-2d watermarking: Embedding messages in 3d meshes and
  extracting them from 2d renderings.
\newblock In {\em Proceedings of the IEEE/CVF Conference on Computer Vision and
  Pattern Recognition}, pages 10031--10040, 2022.

\bibitem{qi2017pointnet++}
Charles~Ruizhongtai Qi, Li~Yi, Hao Su, and Leonidas~J Guibas.
\newblock Pointnet++: Deep hierarchical feature learning on point sets in a
  metric space.
\newblock {\em Advances in neural information processing systems}, 30, 2017.

\bibitem{guo2021pct}
Meng-Hao Guo, Jun-Xiong Cai, Zheng-Ning Liu, Tai-Jiang Mu, Ralph~R Martin, and
  Shi-Min Hu.
\newblock Pct: Point cloud transformer.
\newblock {\em Computational Visual Media}, 7(2):187--199, 2021.

\bibitem{lahav2020meshwalker}
Ayellet Lahav, Alon~Tal.
\newblock Meshwalker: Deep mesh understanding by random walks.
\newblock {\em ACM Transactions on Graphics (TOG)}, 39(6):1--13, 2020.

\bibitem{li2022backdoor}
Yiming Li, Yong Jiang, Zhifeng Li, and Shu-Tao Xia.
\newblock Backdoor learning: A survey.
\newblock {\em IEEE Transactions on Neural Networks and Learning Systems},
  2022.

\bibitem{zhao2020clean}
Shihao Zhao, Xingjun Ma, Xiang Zheng, James Bailey, Jingjing Chen, and Yu-Gang
  Jiang.
\newblock Clean-label backdoor attacks on video recognition models.
\newblock In {\em Proceedings of the IEEE/CVF Conference on Computer Vision and
  Pattern Recognition}, pages 14443--14452, 2020.

\bibitem{chen2022neighboring}
Liang Chen, Qibiao Peng, Jintang Li, Yang Liu, Jiawei Chen, Yong Li, and Zibin
  Zheng.
\newblock Neighboring backdoor attacks on graph convolutional network.
\newblock {\em arXiv preprint arXiv:2201.06202}, 2022.

\bibitem{tian2021poisoning}
Guiyu Tian, Wenhao Jiang, Wei Liu, and Yadong Mu.
\newblock Poisoning morphnet for clean-label backdoor attack to point clouds.
\newblock {\em arXiv preprint arXiv:2105.04839}, 2021.

\bibitem{chen2020end}
Bo~Chen, Alvaro Parra, Jiewei Cao, Nan Li, and Tat-Jun Chin.
\newblock End-to-end learnable geometric vision by backpropagating pnp
  optimization.
\newblock In {\em Proceedings of the IEEE/CVF Conference on Computer Vision and
  Pattern Recognition}, pages 8100--8109, 2020.

\bibitem{gao2003complete}
Xiao-Shan Gao, Xiao-Rong Hou, Jianliang Tang, and Hang-Fei Cheng.
\newblock Complete solution classification for the perspective-three-point
  problem.
\newblock {\em IEEE transactions on pattern analysis and machine intelligence},
  25(8):930--943, 2003.

\bibitem{lepetit2009epnp}
Vincent Moreno-Noguer Lepetit.
\newblock Epnp: An accurate o (n) solution to the pnp problem.
\newblock {\em International journal of computer vision}, 81(2):155--166, 2009.

\bibitem{li2018bundle}
Yanyan Li, Shiyue Fan, Yanbiao Sun, Qiang Wang, and Shanlin Sun.
\newblock Bundle adjustment method using sparse bfgs solution.
\newblock {\em Remote Sensing Letters}, 9(8):789--798, 2018.

\bibitem{besl1992method}
Neil~D Besl, Paul J~McKay.
\newblock Method for registration of 3-d shapes.
\newblock In {\em Sensor fusion IV: control paradigms and data structures},
  volume 1611, pages 586--606. Spie, 1992.

\bibitem{lian2011shape}
Z~Lian, A~Godil, B~Bustos, M~Daoudi, J~Hermans, S~Kawamura, Y~Kurita, G~Lavoua,
  P~Dp Suetens, et~al.
\newblock Shape retrieval on non-rigid 3d watertight meshes.
\newblock In {\em Eurographics workshop on 3d object retrieval (3DOR)}.
  Citeseer, 2011.

\bibitem{feng2019meshnet}
Yutong Feng, Yifan Feng, Haoxuan You, Xibin Zhao, and Yue Gao.
\newblock Meshnet: Mesh neural network for 3d shape representation.
\newblock In {\em Proceedings of the AAAI Conference on Artificial
  Intelligence}, volume~33, pages 8279--8286, 2019.

\bibitem{wu20153d}
Zhirong Wu, Shuran Song, Aditya Khosla, Fisher Yu, Linguang Zhang, Xiaoou Tang,
  and Jianxiong Xiao.
\newblock 3d shapenets: A deep representation for volumetric shapes.
\newblock In {\em Proceedings of the IEEE conference on computer vision and
  pattern recognition}, pages 1912--1920, 2015.

\bibitem{xiang2022detecting}
Zhen Xiang, David~J Miller, Siheng Chen, Xi~Li, and George Kesidis.
\newblock Detecting backdoor attacks against point cloud classifiers.
\newblock In {\em ICASSP 2022-2022 IEEE International Conference on Acoustics,
  Speech and Signal Processing (ICASSP)}, pages 3159--3163. IEEE, 2022.

\bibitem{chou2020sentinet}
Florian~Pellegrino Chou, Edward~Tramer.
\newblock Sentinet: Detecting localized universal attacks against deep learning
  systems.
\newblock In {\em 2020 IEEE Security and Privacy Workshops (SPW)}, pages
  48--54. IEEE, 2020.

\bibitem{huang2019neuroninspect}
Moustafa~Srivastava Huang, Xijie~Alzantot.
\newblock Neuroninspect: Detecting backdoors in neural networks via output
  explanations.
\newblock {\em arXiv preprint arXiv:1911.07399}, 2019.

\bibitem{zheng2019pointcloud}
Tianhang Zheng, Changyou Chen, Junsong Yuan, Bo~Li, and Kui Ren.
\newblock Pointcloud saliency maps.
\newblock In {\em Proceedings of the IEEE/CVF International Conference on
  Computer Vision}, pages 1598--1606, 2019.

\end{thebibliography}

%






\end{document}